%% file: main.tex
\documentclass[10pt]{article}
\newif\ifarxiv \arxivtrue
\ifarxiv \usepackage[preprint]{tmlr/tmlr} \else \usepackage{tmlr/tmlr} \fi
\usepackage[utf8]{inputenc}
\usepackage[T1]{fontenc}

\usepackage{hyperref}
\usepackage{url}
\usepackage{array}
\usepackage{booktabs}
\usepackage{longtable}
\usepackage{amsfonts,amsmath,amssymb,amsthm}
\usepackage{graphicx}
\usepackage{microtype}
\usepackage{xcolor}

\newtheorem{proposition}{Proposition}

\newcommand{\lift}{\mathrm{lift}}

\title{Persistence Floors: What Copying the Last Frame Already Scores in World-Model Evaluation}

\ifarxiv
\author{\name Michael Surazhsky \email surazhsk@post.bgu.ac.il \\
        \addr Cyber@BGU, Ben-Gurion University of the Negev
        \AND \name Yuval Elovici \email elovici@bgu.ac.il \\
        \addr Cyber@BGU, Ben-Gurion University of the Negev
        \AND \name Ronen Brafman \email brafman@bgu.ac.il \\
        \addr Department of Computer Science, Ben-Gurion University of the Negev}
\else
\author{\name Anonymous authors \\ \addr Paper under double-blind review}
\fi

\def\month{MM}
\def\year{YYYY}
\def\openreview{\url{https://openreview.net/forum?id=XXXX}}

\begin{document}

\maketitle

\begin{abstract}
World models are routinely evaluated by how closely their predictions match the future: latent
error, variance explained, LPIPS and SSIM of decoded frames. Every such number has a trivial
reference that is rarely reported, which is the score obtained by copying the most recent
observation forward. We measure this \emph{persistence floor} on four released world models:
DINO-WM, an independent retrain of it, JEPA-WM and LeWM. Each is scored with its own code, on its own
validation split, with prediction and floor computed in one pass. The floor is high, and copying
alone clears an absolute $0.5$ threshold in 28 of 38 (model, environment, read-out) cells. Within a
single checkpoint, the choice among three read-outs of the same visual latent moves the floor by
$0.55$ to $1.37$, depending on model and environment. Across three independently trained models on
one encoder family, the floors agree to within $0.005$ when read per dimension and $0.03$ when read
per patch, so the floor is fixed by the encoder, the data and the read-out before any predictor is
trained. We report
$\lift = (v - \mathrm{floor})/(\mathrm{ceiling} - \mathrm{floor})$, which for variance explained is
the mean-square-error skill score against persistence. We prove that lift is invariant to the
read-out's normalization and confirm it on public checkpoints: between two read-outs the raw score
of an untrained predictor moves by up to $1.86$, while its lift moves by less than $2\times10^{-4}$.
We also place the floor beside published tables. DINO-WM's own LPIPS and SSIM beat copying by
a wide margin, while all four PushT LPIPS baselines in the same table score worse than copying the
previous frame.
\end{abstract}

\section{Introduction}
\label{sec:intro}

A latent world model is trained to predict its own future representations, and the natural way to
report its quality is to say how close those predictions come. Common choices are the coefficient of
determination, latent mean-squared error, cosine similarity, and perceptual distance between decoded
predictions and the true frames. A value near the good end of the scale reads as success.

On its own such a value establishes little, because world models are trained and evaluated at short
horizons and dense sampling, where consecutive observations are nearly identical. A model that has
learned nothing about dynamics and only carries the last observation forward already scores well. We
call that score the \emph{persistence floor}. We first met it in a navigation world model of our own,
where a predictor that explained $97.94\%$ of the variance of its target was beaten by copying the
previous representation, which explained $99.62\%$. The model's squared error was $5.4$ times the
copier's, and an absolute threshold on the score could not have caught it
(Appendix~\ref{app:inhouse}).

Those runs come from one unreleased codebase. This paper therefore measures the floor on the public
checkpoints the field builds on, with their authors' code and validation splits, and asks three
questions.
\begin{enumerate}
  \item \textbf{How high is the floor, and what sets it?} (Section~\ref{sec:floor}.) Copying clears
  an absolute $0.5$ in 28 of 38 cells. The floor is set by the encoder, the data and the read-out.
  Three independently trained patch-grid models agree on the per-dimension floor to within $0.005$,
  while within any one of them the read-out moves the floor by up to $1.37$.
  \item \textbf{Does a better threshold fix it?} (Section~\ref{sec:invariance}.) No threshold on the
  raw score does, because two read-outs of one untrained predictor's outputs differ in raw score by
  up to $1.86$, and their lifts by less than $2\times10^{-4}$. We prove this invariance and show it
  on all 18 predictor cells, trained and untrained.
  \item \textbf{Does it change how any published number reads?} (Section~\ref{sec:typea}.)
  DINO-WM's own LPIPS and SSIM beat copying, at lift $+0.57$ to $+0.93$. The same tables also rank
  baselines that score below the copier, and on PushT that is all four LPIPS baselines.
\end{enumerate}
Section~\ref{sec:init} shows that an untrained residual predictor already clears $0.5$ on one
read-out, with a squared error $8.5$ to $32.4$ times the copier's. Section~\ref{sec:proprio} reports a trained, released
model that clears $0.95$ on one read-out and still loses to copying. Section~\ref{sec:survey} shows
that 11 of 14 recent world-model papers that report a prediction-quality number report nothing to
read it against.

\paragraph{Relation to prior work.} The statistic is established. For variance explained, lift is the
mean-square-error skill score against a persistence reference, which forecast verification has used
for decades \citep{murphy1988skill,murphy1992climatology}. The $R^2$ of Eq.~\eqref{eq:r2} is the same
score against the mean, known in hydrology as the Nash--Sutcliffe efficiency \citep{nash1970river}.
Our contribution is to bring the persistence reference into world-model evaluation, where it depends
on a learned encoder and on the read-out, so it cannot be looked up and must be measured on every run.
The released models we examine beat their floors on every visual read-out, and we report that
alongside the cases that fall below.

\section{The persistence floor and lift}
\label{sec:background}

\paragraph{Floor.} A latent world model predicts the encoder's next representation from the
representations before it and the action taken. Let $z_t$ be that target, the representation at step
$t$, and $\hat z_t$ the model's prediction of it; the index $t$ is dropped when only one step is in
play. A read-out cuts the predictions into rows, each a target vector with its prediction, for
example one row per image patch (Section~\ref{sec:setup} defines the four we use). $\mathbb{E}$ averages over the rows, $\lVert\cdot\rVert^2$ sums the squares of a
row's coordinates, and $\bar z = \mathbb{E}\, z$ is the batch mean of the target, taken coordinate by
coordinate. Score $\hat z$ by the coefficient of determination
\begin{equation}
  R^2 \;=\; 1 - \frac{A}{S}, \qquad A = \mathbb{E}\lVert \hat z - z \rVert^2, \qquad
  S = \mathbb{E}\lVert z - \bar z \rVert^2 .
  \label{eq:r2}
\end{equation}
$A$ is the model's mean squared error. $S$ has the same form with $\hat z$ replaced by $\bar z$: it
is the error of a constant predictor that answers the batch mean for every row, which is also the
total variance of the target. Both sum over all coordinates before the ratio is taken, so $R^2$ is
one number per read-out, not an average of per-coordinate scores. It is the fraction of the constant
predictor's error that the model removes: $1$ is perfect, $0$ is no better than the batch mean, and
there is no lower bound. The persistence floor is the same estimator on the same targets with the
previous representation $z_{t-1}$ in place of $\hat z_t$. Writing
$B = \mathbb{E}\lVert z_{t-1} - z_t\rVert^2$ for the copier's squared error, it is $1 - B/S$. If the
target is wide-sense stationary over the batch, with lag-1 autocorrelation
$\rho = \mathbb{E}\langle z_t - \bar z,\, z_{t-1} - \bar z\rangle / S$, then $B = 2S(1-\rho)$ and
\begin{equation}
  \mathrm{floor} = 2\rho - 1 .
  \label{eq:floor}
\end{equation}
No property of the predictor appears in Eq.~\eqref{eq:floor}. The floor is a property of the embedded
data and the sampling rate, and it approaches $1$ as $\rho \to 1$, which is the regime of dense
sampling and one-step prediction. All three public codebases we examine predict one step ahead,
where a step spans five simulator frames (Section~\ref{sec:setup}), and so does our own. Appendix~\ref{app:horizon} gives the
floor at longer lags.

\paragraph{Lift.} We judge a metric by
\begin{equation}
  \lift = \frac{v - \mathrm{floor}}{\mathrm{ceiling} - \mathrm{floor}},
  \label{eq:lift}
\end{equation}
where $v$ is the model's score on a metric, $\mathrm{floor}$ is the copier's score on the same rows,
and $\mathrm{ceiling}$ is the score of a perfect prediction. Lift is $0$ at the copier and $1$ at the
ceiling. For $R^2$ with ceiling $1$, Eq.~\eqref{eq:lift} reduces to $1 - A/B$, the MSE skill score
against persistence \citep{murphy1988skill,murphy1992climatology}. The three scores thus compare
three predictors pairwise: $R^2 = 1 - A/S$ sets the model against the batch mean,
$\mathrm{floor} = 1 - B/S$ the copier against the batch mean, and $\lift = 1 - A/B$ the model
against the copier.
For LPIPS the ceiling is $0$ and for SSIM it is $1$. Eq.~\eqref{eq:lift} applies unchanged: for LPIPS,
where lower is better, numerator and denominator change sign together. We apply three rules. First, lift is never clamped. A negative lift
is a result, and for $R^2$ the quantity $1-\lift$ is the ratio of the model's squared error to the
copier's. Second, a metric with no declared ceiling carries no threshold. Third, when the ceiling
sits on the wrong side of the floor, or the two are closer than $10^{-4}$ of their magnitude, the
implementation returns \texttt{n/a} instead of a number.

\paragraph{Two kinds of evidence.} \emph{Threshold measurements} ask what an absolute threshold on a
latent-prediction score would conclude. No paper we examine publishes such a score for these
checkpoints, so these measurements establish where the floor sits and cannot show that a published
number misleads. \emph{Comparisons with published numbers} place a floor beside a number that a paper
printed. We count the two kinds separately throughout.

\section{Setup}
\label{sec:setup}

\paragraph{Models.} We evaluate four released world models, on the environments for which weights
exist and which need no deformable-body simulator.
\textbf{DINO-WM} \citep{zhou2024dinowm}: the authors' checkpoints for PointMaze, PushT and Wall.
The model has a frozen DINOv2 ViT-S/14 encoder \citep{oquab2024dinov2}, a patch grid of $196$ tokens
(the encoder receives $196$\,px: the configuration and the paper's hyperparameter table give $224$,
but the model resizes a second time, as the paper's appendix also states), a ViT predictor, and a
\emph{proprio encoder}: one learned linear layer that embeds the agent's low-dimensional state
vector. Its output is copied to every patch and concatenated to each patch token along the channel
axis, and the predictor predicts it along with the visual channels. Actions enter the same way,
through an action encoder.
\textbf{Meta DINO-WM} and \textbf{JEPA-WM} \citep{terver2026drives}: Meta DINO-WM is an independent
retrain of DINO-WM, and JEPA-WM is trained with the JEPA-WM recipe. The recipe's authors released
both, trained at $224$\,px ($256$ patches) on DINO-WM's data. JEPA-WM's predictor is conditioned
through AdaLN \citep{peebles2023dit}.
\textbf{LeWM} \citep{lewm2026}: a ViT-tiny encoder whose CLS token is projected to a single
$192$-dimensional latent, so the model has no patch grid. It takes pixels only, with no
proprioceptive input. All four predict one step ahead, and a
step spans five simulator frames.

\paragraph{Read-outs.} The same predictions are scored four ways. Let a latent be a grid of $P$
patches by $D$ channels. \emph{Pooled} averages over patches first and then scores the $D$-vector,
which changes the squared error itself. In the \emph{per-patch} read-out every (window, patch) is a
row of $D$ channels, and $S$ is the spread about each channel's mean. In the \emph{per-dim} read-out
every (window, channel) is a row of $P$ patches, and $S$ is the spread about each patch's mean. So
per-patch $S$ includes how much patches differ from one another, and per-dim $S$ how much channels
differ (Section~\ref{sec:floor}). \emph{Proprio} scores the predicted proprioceptive embedding.
Per-patch and per-dim arrange the same squared errors into different rows. Their $A$ and $B$
therefore agree up to one common factor, the ratio of the two row counts, and beyond that factor
the two read-outs differ only in $S$. LeWM has one token, so
its two read-outs are the token as a whole and its $192$ dimensions one at a time.

\paragraph{Protocol.} Each model is scored with its own released code, loader and validation split
(DINO-WM seed 42, JEPA-WM seed 234, LeWM seed 3072), on $n=1024$ validation windows. Prediction,
copier and target come from \emph{one forward pass}, so value and floor share the estimator, the
targets and the windows by construction. Three arms are scored per model: the released
\emph{checkpoint}; \emph{copy-forward}, which substitutes $z_{t-1}$ for the prediction; and
\emph{random}, the same predictor architecture at initialisation with the trained encoder. Vendor
commits are pinned (Appendix~\ref{app:protocol}).

\paragraph{Copier control.} The copy-forward arm sends the copier through the same scoring path as a
model, so its lift must be exactly $0$. It is, on all 50 pairs: the 38 cells of
Section~\ref{sec:floor} and, for the pixel metrics of Section~\ref{sec:typea}, two metrics on three
environments, each over all windows and over the half with the most motion. This confirms that value
and floor come from one estimator on the same rows. It cannot show that the copier is the right
frame, because the arm and the floor read the same array. That rests on each model's own code: the
copier is the tensor the authors' predictor takes as input, and the target is the tensor their
training loss compares the prediction with.

\paragraph{Reproduction checks.} A floor is placed beside a published number only after the harness
reproduces that number. Table~\ref{tab:gates} lists the checks. DINO-WM's pixel metrics are
reproduced on both published environments. JEPA-WM and Meta DINO-WM are checked on PushT planning
success, and both land inside a band of $\pm10$ points that we fixed before the run. LeWM reports
PushT planning success only in an appendix table; our result lies inside the same band, which for
LeWM was not fixed in advance.

\begin{table}[t]
  \centering
  \caption{Reproduction checks: the published number each harness must return before any floor from
  it is quoted. Planning rows give success in percent; the parenthesis after a published value is the
  spread its paper reports over training seeds. DINO-WM's paper states no variance, so we set its
  band ourselves: LPIPS within a factor of $2$, SSIM within $\pm0.01$.}
  \label{tab:gates}
  \small
  \begin{tabular}{@{}llll@{}}
    \toprule
    Target & Number & Published & Ours \\
    \midrule
    DINO-WM & PushT LPIPS / SSIM, one step & $0.007$ / $0.985$ & $0.0067$ / $0.9867$ \\
    DINO-WM & Wall LPIPS / SSIM, one step & $0.0016$ / $0.997$ & $0.0022$ / $0.9972$ \\
    JEPA-WM & PushT planning, CEM-L2 & $70.2$ ($2.8$) & $63.9$ \\
    Meta DINO-WM & PushT planning, CEM-L2 & $66.0$ ($4.7$) & $62.2$ \\
    LeWM & PushT planning, CEM & $96.0$ ($2.83$) & $92$ ($46/50$) \\
    \bottomrule
  \end{tabular}
\end{table}

\paragraph{Intervals.} Windows from one episode are correlated, so we resample whole episodes
\citep{field2007clustered}, $10{,}000$ times, and recompute value and floor on each resample.
The validation splits hold 200 (PointMaze), 191--192 (Wall) and 21 (PushT) episodes for the patch-grid
models and 997 for LeWM, so PushT intervals rest on the fewest episodes. A lift interval that lies
entirely above $0$ means the predictor beats copying; one entirely below $0$ means it loses to
copying. We compute 112 lifts for anything other than the copier: 76 on latents, a trained and an
untrained predictor in each of the 38 cells, and 36 on pixels (Section~\ref{sec:typea}). Every
interval lies on one side of $0$ except one, which belongs to the \emph{decoded copy}, the previous
latent passed through the decoder: its SSIM on the high-motion half of PushT. Before
computing any interval we pre-registered which ones would lie on one side of $0$, and all of those
do. One expectation was wrong: DINO-WM's DINO CLS baseline (Section~\ref{sec:typea}), which we expected to be
indistinguishable from copying, is distinguishably worse. As a control, the copier scored as a model
has lift exactly $0$ in every resample, on all 50 pairs. The published values of
Section~\ref{sec:typea} state no variance, so for them only our floor is resampled.

\section{How high the floor is, and what sets it}
\label{sec:floor}

\begin{table}[t]
  \centering
  \caption{DINO-WM's released checkpoints on each environment and read-out. \emph{Value}: $R^2$ of
  the prediction. \emph{Floor}: $R^2$ of copying the previous representation, on the same rows.
  \emph{Lift}, Eq.~\eqref{eq:lift}: $0$ is as good as copying, $1$ is perfect, below $0$ is worse than
  copying. \emph{Random init}: the same predictor architecture with untrained weights, scored against
  the same floor. \emph{95\% interval}: on the trained lift, from resampling whole episodes
  (Section~\ref{sec:setup}). Every interval, trained and untrained, lies on one side of $0$, so each
  predictor is distinguishably better or worse than copying. One pass per row, $n=1024$ windows.
  Per-patch and per-dim lifts are equal, as Section~\ref{sec:invariance} proves.}
  \label{tab:public}
  \small
  \input{table_public.tex}
\end{table}

\paragraph{Level.} Table~\ref{tab:public} gives DINO-WM's released checkpoints on every read-out.
Across all four models, the copier, which has no parameters, clears an absolute $0.5$ threshold in
28 of the 38 (model, environment, read-out) cells. Read per-dim on DINO-WM, it scores between
$0.9458$ and $0.9844$. Every trained checkpoint beats the copier on every visual read-out, and
DINO-WM's lift runs from $+0.74$ to $+1.00$ across environments and read-outs. These models
therefore pass, but a raw score of $0.95$ is not by itself evidence that a model has learned
dynamics, because copying reaches it too.

\paragraph{Dependence on the read-out.} Within one checkpoint and one pass, the floor of the visual
latent spans $-0.3312$ to $+0.9844$ on PointMaze, $0.3505$ to $0.9458$ on PushT, and $0.1564$ to
$0.9615$ on Wall, over the pooled, per-patch and per-dim read-outs. Proprio scores a different
target and is left out of these ranges. Over the nine (model, environment) pairs of the three
patch-grid models, the spread ranges from $0.55$ to $1.37$. The cause is the denominator
$S$. Per-patch rows mix patches, so $S$ includes the variance between patches, which differ from one
another by position. Per-dim rows mix channels, so $S$ includes the variance between channels. The
copier reproduces both exactly, so $S$ grows while the copier's summed error does not, and the floor
rises. Between these two read-outs, which score the same errors, the floor moves by $0.02$ to
$0.09$. Pooling
averages the patches before scoring, which changes the error and removes the between-patch variance
from $S$, and the floor
falls. On PointMaze it falls below zero, where copying is worse than emitting the batch mean. An
absolute threshold on this checkpoint therefore means something different for each read-out, and
most papers do not state the read-out.

\begin{table}[t]
  \centering
  \caption{Persistence floor per read-out for three independently trained patch-grid models on the
  same data. Zhou's DINO-WM runs at $196$ patches; Meta's two at $256$. Per-patch and per-dim floors
  agree across models. Proprioceptive floors differ on PointMaze and PushT, because each model
  learns its own proprio encoder.}
  \label{tab:crossmodel}
  \small
  \input{table_crossmodel.tex}
\end{table}

\paragraph{Agreement across models.} Table~\ref{tab:crossmodel} places the floors of Zhou's DINO-WM,
Meta's retrain and JEPA-WM side by side. These are three training runs by two teams, at two
resolutions and with two predictor architectures, on one frozen encoder family and one dataset. Read
per-dim, their floors agree to within $0.0042$ in every environment, and per-patch to within
$0.0280$. Pooled floors differ by up to $0.1028$. An average over $196$ patches and an average over
$256$ are different quantities, but resolution is not the whole cause: the two $256$-patch models
also differ, by up to $0.0398$. The floor contains no predictor, so the per-patch and per-dim agreement
says that two codebases at two resolutions measure the same visual floor: it is set by encoder, data
and read-out, and can be computed before any predictor is trained. This holds for the frozen visual
channel only. Proprioceptive floors differ across
the three models by up to $0.4917$, because each model learns its own proprio encoder, and a learned
target has its own floor.

\paragraph{A single-token model.} LeWM's one $192$-d token has no patch axis, and the read-out barely
moves its floor: $0.8103$ over the token and $0.8108$ per dimension, a spread of $0.0005$. On the
same task and frameskip, DINO-WM's floor moves by $0.0921$ between per-patch and per-dim and by
$0.5954$ over its three visual read-outs (Table~\ref{tab:lewm}). Even the $192$
single-dimension floors lie between $0.6763$ and $0.8677$. The read-out therefore moves the floor
only when the latent has an axis, such as patches, along which the means differ. The level of the
floor does not depend on such an axis, since copying still clears $0.5$ on both LeWM read-outs. LeWM itself beats its
floor, with value $0.9919$ and lift $+0.96$.

\section{Lift is invariant to the read-out's normalization}
\label{sec:invariance}

A natural response to Section~\ref{sec:floor} is to set a threshold per read-out, for example $0.95$
for per-dim and $0.5$ for pooled. This section shows why that fails and why lift needs no such
adjustment.

\begin{proposition}
Let two read-outs score the same predictions and targets, so that their squared errors agree up to a
constant $c > 0$: $A_2 = cA_1$ for the model and $B_2 = cB_1$ for the copier. Let their normalising
spreads $S_1$ and $S_2$ be arbitrary. Their raw scores $1 - A_i/S_i$ and floors $1 - B_i/S_i$
differ in general, but with ceiling $1$ their lifts are equal: $\lift_1 = \lift_2 = 1 - A_1/B_1$.
\end{proposition}
\begin{proof}
$\lift_i = \big((1 - A_i/S_i) - (1 - B_i/S_i)\big)/\big(1 - (1 - B_i/S_i)\big) = (B_i - A_i)/B_i$,
independent of $S_i$, and $A_2/B_2 = A_1/B_1$.
\end{proof}

\begin{figure}[t]
  \centering
  \includegraphics[width=\textwidth]{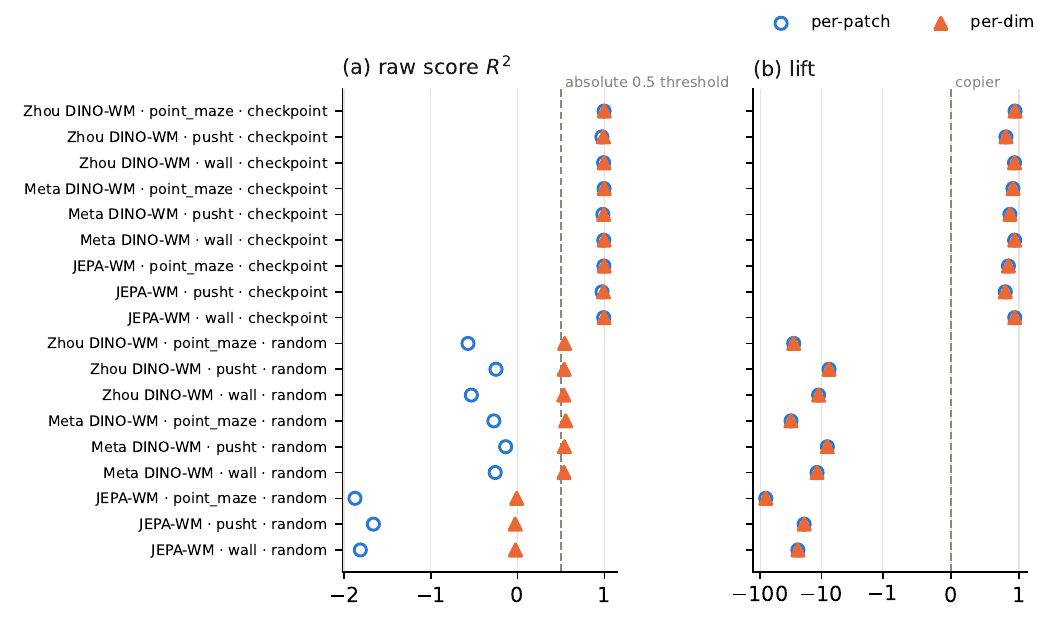}
  \caption{The same predictions, read two ways. \textbf{(a)} Raw $R^2$ per-patch (circles) and
  per-dim (triangles) for every predictor cell; the dashed line is an absolute $0.5$ threshold. On
  the untrained DINO-WM rows the two read-outs land on opposite sides of it. \textbf{(b)} Lift for the
  same cells (symmetric-log axis); the dashed line is the copier. The markers coincide in every row,
  to within $1.9\times10^{-4}$. Pooled is omitted because it changes the error itself
  (Section~\ref{sec:invariance}).}
  \label{fig:invariance}
\end{figure}

Per-patch and per-dim are such a pair, with $c$ the ratio of their row counts
(Section~\ref{sec:setup}), so the proposition can be checked on data. Figure~\ref{fig:invariance} plots all 18 predictor
cells (three models, three environments, trained and untrained). The two raw scores differ by up to
$1.8639$ on the untrained predictors and by up to $0.0176$ on the trained checkpoints. On the six
untrained DINO-WM rows, Zhou's and Meta's, they fall on opposite sides of the
$0.5$ threshold. DINO-WM's untrained PointMaze predictor, for example, reads $+0.5424$ per-dim and
$-0.5715$ per-patch, and both of its lifts read $-28.39$. Across all 18 cells the two lifts differ
by at most $1.9\times10^{-4}$, on a lift of $-81.17$, and on the trained checkpoints by at most
$2.2\times10^{-6}$; gaps of this size are consistent with the rounding of single-precision
arithmetic. No single threshold on the raw score can be right for both read-outs,
whereas lift needs no adjustment, because the normalization cancels.

Pooling falls outside the proposition. It averages over patches before computing the error, so $A$
itself changes. Pooled lift differs from per-patch lift (for DINO-WM on PushT, $+0.74$ against
$+0.81$) because pooling poses a different prediction problem. Which error to report remains a
scientific choice. Once lift is used, how that error is normalised no longer affects the result.

\section{What an untrained predictor scores}
\label{sec:init}

The random-init columns of Table~\ref{tab:public} show what a predictor scores before any training.
DINO-WM's predictor is a ViT whose blocks add their output to a residual stream, followed by a final
LayerNorm, so at initialisation its input has a direct path to its output. Read
per-dim, untrained DINO-WM predictors reach $+0.5424$, $+0.5374$ and $+0.5328$ on the three
environments, clearing $0.5$ on all three. Meta's retrain of the same architecture reproduces this,
at $+0.5543$, $+0.5400$ and $+0.5348$. None of these predictors is close to the copier. Their lifts
are $-7.54$ to $-31.42$, so their squared error is $8.5$ to $32.4$ times the copier's, and read
per-patch the same predictions score below zero, from $-0.1374$ to $-0.5715$.

The per-dim score is therefore no evidence of copying, and a constant prediction shows what it
rewards. Predict, for every patch of every window, the batch mean of each channel. The summed
squared error of this constant equals the per-patch spread, so its per-dim score follows from the two
floors as $1 - (1 - \mathrm{floor}_{\text{per-dim}})/(1 - \mathrm{floor}_{\text{per-patch}})$. On
these six (model, environment) pairs it is $0.60$ to $0.71$, above the untrained predictor in every
pair. Per-dim $S$ includes the variance between channels (Section~\ref{sec:floor}), and any output
that keeps the channels' static offsets is credited with it. A path from input to output is one way
to keep them; we did not test what else that path carries. The same untrained DINO-WM predictor
reads $-118.8873$ pooled on PointMaze, where pooling has removed the between-patch variance from $S$.

JEPA-WM's predictor does not behave this way: at initialisation its per-dim score is
$-0.0098$ to $-0.0272$. It is conditioned through AdaLN, with randomly initialised projections
between its input and its output, which is the likely reason. LeWM's untrained predictor reads
$-0.0415$ over its token.

Two consequences follow. First, an absolute threshold can be passed by a network that has learned
nothing: $0.5$ is cleared, on one read-out, by predictors whose error is many times the copier's.
Second, the absolute score of an untrained model depends on an architectural detail, whether the
input has a path to the output, and on the read-out, and neither concerns dynamics. Lift places
every one of these predictors far below zero on every read-out. The case lift exists to catch is the
trained predictor that has learned the identity and sits at lift zero;
Appendix~\ref{app:collapse} explains why anti-collapse regularisers do not prevent it.

\section{Published numbers against the floor}
\label{sec:typea}

The measurements so far are threshold measurements. This section places the floor beside numbers
that a paper printed.

\paragraph{Which numbers, and what produced them.} DINO-WM reports LPIPS
\citep{zhang2018lpips} and SSIM \citep{wang2004ssim} of decoded predictions in its Tables~5 and~6,
for its own model and four baselines: R3M \citep{nair2022r3m}, ResNet, DINO CLS, and AVDC
\citep{ko2024avdc}. Table~7 of the same paper gives the same metrics for DINO-WM retrained on subsets
of the PushT data. The paper states no horizon, split or sample size for
these numbers, and its text places them beside a figure of open-loop rollouts. The released code
shows which evaluation produced them. The open-loop evaluation never calls the image-metric function.
The only code path that emits image LPIPS and SSIM is the validation loop of training, which decodes
a \emph{teacher-forced, one-step} prediction (one step, spanning five frames, from ground-truth context) and
compares it with the true next frame. We reproduce that evaluation with the authors' own metric
function (LPIPS with a VGG backbone, Gaussian SSIM) on their validation split, with $n=1024$ windows.
In the same call we score the copier, which emits the previous ground-truth frame; the
\emph{decoded copy}, the previous latent passed through the decoder; and the \emph{decoded target},
the true next latent passed through the decoder, which is the best score any predictor can reach
through this decoder. Each is scored over all windows and over the half with the most
frame-to-frame motion. The reproduction check passes on both published environments
(Table~\ref{tab:gates}).

\begin{table}[t]
  \centering
  \caption{DINO-WM pixel metrics, teacher-forced one step. \emph{Copy floor}: the previous frame,
  scored by the authors' metric function in the same pass. \emph{Decoded copy}: the previous latent
  through the decoder. \emph{Lift of published}: the lift of DINO-WM's published value, with its 95\%
  interval; only our floor is resampled, since the published value states no variance. \emph{Lift,
  ours, high motion}: our reproduction's lift on the half of windows with the most frame-to-frame
  motion. PointMaze has no published row.}
  \label{tab:typea}
  \small
  \setlength{\tabcolsep}{4pt}
  \input{table_type_a.tex}
\end{table}

\paragraph{DINO-WM's own rows.} DINO-WM's published values beat the copy floor by a wide margin
(Table~\ref{tab:typea}): lift $+0.80$ (LPIPS) and $+0.57$ (SSIM) on PushT, and $+0.93$ and $+0.78$ on
Wall. None of its four published numbers falls below the floor. We checked two alternative
explanations. A static background does not account for the score, since on the half of windows with
the most motion our reproduction's lift is higher still ($+0.83$ and $+0.68$ on PushT). Decoder blur
does not account for the gap either, since the decoded copy scores slightly worse than the raw
copier (11 of its 12 lift intervals lie below $0$), and the model beats both.

\begin{table}[t]
  \centering
  \caption{DINO-WM's published pixel metrics (its Tables~5--7; one step, teacher-forced) with the
  copier added as a row. Each cell: published value (lift against copying the previous frame). Bold:
  at or below the copier. The copier row gives the floor, whose lift is $0$ by definition. Every lift's
  95\% interval lies on one side of $0$; only our floor is resampled. AVDC was evaluated in its own
  codebase. The 200-trajectory retrain is published for PushT only.}
  \label{tab:baselines}
  \small
  \input{table_baselines.tex}
\end{table}

\paragraph{The baseline rows.} The floor is computed from the data alone, so it applies to every row
of DINO-WM's tables, and Table~\ref{tab:baselines} adds it as one more row. On PushT, all four LPIPS
baselines score worse than copying the previous frame. The one closest to the floor, DINO CLS
($0.039$ against $0.0352$, lift $-0.11$), has lift interval $[-0.176, -0.046]$. The three baselines
that share DINO-WM's training code (R3M, ResNet and DINO CLS) have 12 published numbers, two metrics
on two environments each, and 5 of them fall at or below the floor. AVDC, evaluated in its own
codebase, falls below it on all 4 of its numbers. DINO-WM's own smallest-data retrain, on 200
trajectories (its Table~7), is below the floor on both metrics. A reader of DINO-WM's Table~5 sees
five methods ranked by LPIPS. With a copier row, the PushT column shows one method that beats
copying the previous frame and four that do not; on Wall, four of the five beat it.

\paragraph{Caveats.} DINO-WM beats every baseline and the copier, so its ranking stands. What the
copier row adds is a reference point: it shows which of the ranked methods beat copying the previous
frame and which do not. Five caveats bound this claim.
\begin{enumerate}
  \item \emph{Post-hoc.} The prediction we registered before measuring the pixel floor covered only
  DINO-WM's own rows. It was right
  that the PushT values would stay above the floor and wrong about the size of the Wall floor. We did
  not predict that any baseline would fall below the floor; we found that by applying the same floor
  to the rest of the table. The later prediction about intervals (Section~\ref{sec:setup}) was made
  with those results known.
  \item \emph{Protocol.} R3M, ResNet and DINO CLS share DINO-WM's validation loop, so their protocol
  is the one we reproduced. AVDC was evaluated in its own codebase, and its protocol is unverified.
  The rows of DINO-WM's Table~7 are retrains on data subsets whose validation split may differ from
  ours.
  \item \emph{Noise in the published values.} In the released code each published number comes from
  one validation batch, and the paper gives DINO-WM's PushT LPIPS as $0.007$ in its Table~5 and, for
  training on all $18{,}500$ trajectories, $0.005$ in its Table~7. The smallest margin, DINO CLS at
  $0.039$ against $0.0352$, is less than twice the gap between those two values. Its lift interval,
  $[-0.176, -0.046]$, excludes $0$, but it reflects only the
  uncertainty of our floor, which is estimated from the 21 episodes of PushT's validation split:
  $[0.0332, 0.0373]$. The published value's own sampling noise is not stated and not covered. We
  count DINO CLS as below the floor and flag it as the one result that a single noisy validation
  batch could explain.
  \item \emph{Teacher forcing.} The published numbers are one-step and teacher-forced. Whether the
  ordering survives open-loop rollouts, where the copier degrades with horizon, is a separate
  question that we do not test.
  \item \emph{Decoders.} The copier row is an undecoded frame, while every method's prediction
  passes through that method's own decoder, and lift here uses the metric's ceiling ($0$ for LPIPS,
  $1$ for SSIM), so decoder loss counts against the method. For DINO-WM the decoder costs little:
  its decoded target scores LPIPS $0.0045$ and SSIM $0.9952$ on PushT, against the copier's $0.0352$
  and $0.9654$. We have no such measurement for the baselines. A baseline row below the copier
  shows that its decoded prediction is a worse picture of the next frame than the previous frame
  is; it does not separate a weak predictor from a lossy decoder.
\end{enumerate}

\section{A trained model below its floor}
\label{sec:proprio}

In the threshold measurements above, every predictor that clears $0.5$ while losing to the copier is
untrained, with one exception. Meta's released JEPA-WM, whose PushT planning we reproduced, predicts
the Wall proprioceptive embedding with $R^2 = +0.9557$, far above the $0.5$ threshold.
Copying the previous proprio embedding scores $0.9791$, so its lift is $-1.12$, with interval
$[-1.47, -0.83]$ over 191 episodes, and its squared error is more than twice the copier's. Meta's
DINO-WM retrain does worse on the same read-out ($R^2 = -0.5097$ against a floor of $0.9784$), while Zhou's
DINO-WM beats the copier there (lift $+0.82$). The same harness gives proprio lifts of $+0.97$ to
$+0.999$ for both Meta models on PointMaze and PushT, so the proprio read-out works on two of the
three environments. A fault specific to Wall in our harness for Meta's models is not excluded: both
of them fall below the copier there, and the untrained Meta DINO-WM predictor reads
$R^2 = -688976.9$ on this read-out, a value we have not traced.

We have not established the cause. The Wall floor is $0.98$ because the agent moves little in five
frames, so the margin a model must earn is $0.02$. One untested hypothesis is that a proprio loss
summed beside a $384$-dimensional visual loss is too small a share of the training loss to earn it. Wall planning was not reproduced
here; only PushT planning was. If the measurement stands, the case is the kind of failure an
absolute threshold is least able to
catch, a trained, released model whose score reads as excellent and whose prediction is worse than
copying the previous value.

\section{Reporting practice}
\label{sec:survey}

We checked 20 recent world-model papers for whether they report a trivial baseline (copy-last-frame,
persistence, zero-motion or a constant prediction) against which their prediction-quality number can
be read. The coding criterion and per-paper verdicts are in Appendix~\ref{app:survey}. Six report no
prediction-quality number and are judged on task success, which is a legitimate choice. Of the 14 that do
report one, 11 give no floor of any kind, 2 acknowledge an interpretability caveat without measuring a
floor, and 1 measures it.

The eleven include the JEPA-WM recipe paper, which names embedding-space error under unrolling as a
measure that ``track[s] the world model quality, independently of the planning procedure''
\citep[\S5.1]{terver2026drives}, and DINO-WM, whose pixel tables are examined in
Section~\ref{sec:typea}. They also include a platform built for reproducible evaluation, which
reports trajectory-level prediction MSE with no floor \citep[\S4.2]{bounou2026stableworldmodel} and
is where the practice would appear if it were routine. The one paper that measures a floor
forecasts force $0.4$\,s ahead on real-robot data against persistence \citep[Table~16]{latentknow2026}.
Persistence scores $0.23$ and $0.31$ where the model scores $0.43$ and $0.47$, lifts of $0.26$ and
$0.23$.
The practice is routine in neighbouring fields. \citet{villegas2017decomposing} reported a
copy-last-frame baseline overtaking their video predictor in PSNR beyond the first few steps, and
\citet{beck2025naive} argue that machine-learning forecasting still neglects the naive forecast. The
technique exists; what world-model evaluation lacks is the convention of reporting it.

\section{Recommendations}
\label{sec:recs}

\begin{enumerate}
  \item \textbf{Report the floor beside every prediction-quality number}, computed with the same
  estimator on the same targets in the same pass. It costs one extra metric evaluation, since the
  previous observation is already in the batch.
  \item \textbf{Report lift, unclamped}, and put any threshold on lift instead of the raw score. A
  negative lift is a result to report.
  \item \textbf{State the read-out}: over what rows, and about what mean, the score is normalised.
  Under lift the choice of normalization no longer matters (Section~\ref{sec:invariance}), but the
  raw score cannot be read without it.
  \item \textbf{Add a copier row to comparison tables.} It is model-free, so one row serves every
  method in the table (Section~\ref{sec:typea}).
  \item \textbf{State the evaluation path} (open-loop or teacher-forced, the horizon, and the split).
  The copier's strength depends on all three.
  \item \textbf{Report the floor at every horizon evaluated.} The floor at a longer lag cannot be
  inferred from the one-step floor (Appendix~\ref{app:horizon}).
\end{enumerate}
Persistence is the cheapest reference. The optimal linear combination of persistence and the batch
mean, which is the AR(1) predictor, is the natural next one \citep{murphy1992climatology}, and under
the assumptions of Section~\ref{sec:background} it scores above the copier
(Appendix~\ref{app:horizon}). A model that beats persistence but not the AR(1) predictor has learned
less than its lift suggests.

\section{Limitations}
\label{sec:limits}

We measured four models in three simulated environments at one prediction step. We did not evaluate
deformable-object environments, real-robot data, or open-loop horizons. The floor changes with the
lag (Appendix~\ref{app:horizon}) and has to be measured at each horizon, so whether the published
orderings of Section~\ref{sec:typea} survive at longer horizons is untested. The comparison with
published numbers covers one paper. DINO-WM is the one model with released weights, a published
pixel number and code that reveals how the number was made, and the baseline results were found
post-hoc. JEPA-WM and Meta DINO-WM were checked on PushT planning only; Wall and PointMaze planning
were not reproduced. LeWM's planning check uses a band that was not fixed before its run.

Lift can be raised through the encoder. Embedding a signal that changes quickly but is trivially
predictable, such as a copy of the action or a step counter, raises the copier's error without
raising the model's, so lift rises while nothing about the dynamics has been learned. Lift describes
the prediction of the target it is given, and a learned target must be audited separately. A positive
lift says that a prediction number carries information about dynamics; it does not say that the
representation supports planning \citep{rcaux2026}. The survey establishes that floors are rarely
reported; it does not show that any surveyed conclusion is wrong.

\paragraph{Reproducibility.} Every value, floor, lift and interval we measured is generated by a
script from a single store of (value, floor, lift) records, and the text copies it from that
script's output; the numbers of
Appendix~\ref{app:inhouse} come the same way from the in-house run log. The planning successes of
Table~\ref{tab:gates} are the output of the two planning scripts and are not in the store. The
store, the read and planning scripts,
the bootstrap with its per-window scores, the run log and the pinned vendor commits accompany the
paper as supplementary material.

\ifarxiv
\subsubsection*{Acknowledgments}
This work was funded by Cyber@BGU (CBG), Ben-Gurion University of the Negev.
\fi

\bibliography{references}
\bibliographystyle{tmlr/tmlr}

\appendix

\section{Protocol details and declared deviations}
\label{app:protocol}

\paragraph{Pinned code.} \texttt{gaoyuezhou/dino\_wm} at \texttt{0a9492fa} (2025-03-24);
\texttt{facebookresearch/jepa-wms} at \texttt{13cf1d9c} (2026-04-11); \texttt{lucas-maes/le-wm} at
\texttt{8edfeb33} (2026-05-22).

\paragraph{Constants found by running the code.} DINO-WM's configuration states an
image size of $224$, but the model inserts a second resize before the encoder
($224/16 \times 14 = 196$), so the encoder sees $196$\,px and returns $196$ patches. An earlier
floor-only pass encoded at $224$\,px, and none of its values is quoted here. Normalization is
$0.5/0.5$, not ImageNet statistics. DINO-WM predicts one step (\texttt{num\_pred}$=1$), spanning five frames,
with $3$ frames of history on PushT and $1$ on Wall. The paper's hyperparameter table gives $224$
and its App.~A.7 gives $196 \times 196$.

\paragraph{Declared deviations, JEPA-WM.} Decoder heads are off; they are visualisation-only, and
the state head was never released. CEM samples are scored in chunks of $50$ to fit memory, which is
exact. Reads use the $4$-frame training clip and the seed-234 split. PointMaze planning, which needs
MuJoCo~2.1, was not run.

\paragraph{Declared deviations, LeWM.} The validation split is over stride-$1$ windows, so it shares
episodes with training. The random arm re-initialises the predictor and its projection, while the
encoder, projector and action encoder remain trained.

\paragraph{Pre-registration.} Every measurement except the first DINO-WM latent pass was preceded by
a written prediction, recorded before its result was seen. That first pass had none, and we treat it
as exploratory; the Meta retrain reproduces its values to within the cross-model gaps of
Table~\ref{tab:crossmodel}. The baseline results of Section~\ref{sec:typea} were not predicted.

\begin{table}[h]
  \centering
  \caption{LeWM, released PushT checkpoint, one $192$-d token. $n=1024$ windows, one pass.}
  \label{tab:lewm}
  \small
  \input{table_lewm.tex}
\end{table}

\section{Where this started: an in-house model}
\label{app:inhouse}

The measurements that prompted this work come from our own world model for a simulated navigation
task. The agent sees a $32$-ray depth scan and its body state, and the model predicts the next
$27$-dimensional embedding of the scan and the next $5$-dimensional body state, one step ahead, where
a step is one skill of five simulator steps. Ninety-nine checkpoints give 197 score--floor pairs.
Eighteen clear an absolute $0.5$ while predicting worse than copying. The clearest case is a pair of
checkpoints trained identically except for how many pieces the encoder splits the observation into.
With one piece the model scores $0.8687$ over a floor of $0.2791$ (lift $+0.82$). With four pieces it
scores $0.9376$ over a floor of $0.9886$ (lift $-4.45$). The raw score ranks the two in the wrong
order. These runs use unreleased code, and no claim in the main text rests on them.

\begin{figure}[h]
  \centering
  \includegraphics[width=\textwidth]{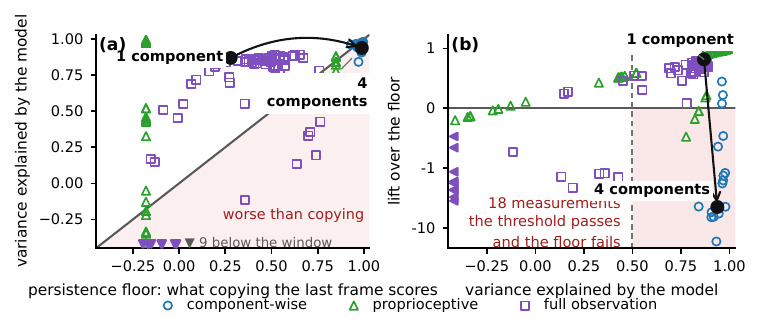}
  \caption{In-house model, all 197 measurements. \textbf{(a)} Score against its own persistence floor.
  \textbf{(b)} Raw score against lift (symmetric-log); the shaded box holds the 18 measurements an
  absolute $0.5$ threshold passes and the floor fails.}
  \label{fig:inhouse}
\end{figure}

\begin{table}[h]
  \centering
  \caption{In-house model, per read-out: pairs below their floor, how many still clear $0.5$, the floor's
  range, and the highest-scoring measurement below its floor.}
  \small
  \input{table1.tex}
\end{table}

\section{The floor as a function of horizon}
\label{app:horizon}

Let $\rho_k = \mathbb{E}\langle z_t - \bar{z},\, z_{t-k} - \bar{z}\rangle/S$ be the autocorrelation
of the target at a lag of $k$ simulator steps. The argument behind Eq.~\eqref{eq:floor} gives
$\mathbb{E}\lVert z_t - z_{t-k} \rVert^2 = 2S(1-\rho_k)$, so copying the representation from
$k$ steps back scores
\begin{equation}
  \mathrm{floor}(k) \;=\; 2\rho_k - 1 .
  \label{eq:floor-k}
\end{equation}
A model evaluated $h$ steps ahead at a frame skip of $f$ simulator steps is read against lag
$k = fh$. The released models of Section~\ref{sec:setup} and our
in-house runs share one setting, $f = 5$ and $h = 1$. The copy floor reaches zero where
$\rho_k = 1/2$; beyond that lag the batch mean is the stronger trivial predictor. Neither is the
strongest simple reference: under the same assumptions the AR(1) predictor
$\bar{z} + \rho_k (z_{t-k} - \bar{z})$ scores $\rho_k^2$, which is never below $0$ and exceeds
$2\rho_k - 1$ for every $\rho_k < 1$, since $\rho_k^2 - (2\rho_k - 1) = (1 - \rho_k)^2$. We keep the copier as the
floor because it fits nothing, while the AR(1) reference fits $\rho_k$; the AR(1) reference is the
stricter choice when one is wanted.

The floor at a longer horizon cannot be inferred from the floor at one step. Version~3 of
\citet[Table 15]{latentknow2026v3} reports force-forecasting floors of $0.65$ and $0.61$ at one step
(learned model $0.72$ and $0.68$) and $0.24$ and $0.31$ at four; the current version keeps only the
longer horizon. Under the stationarity assumed above, and with lags counted in that paper's
prediction steps, they imply $\rho_1 = 0.825$ and $0.805$, and $\rho_4 = 0.620$ and $0.655$. If the
autocorrelation decayed geometrically, $\rho_4$ would be $0.463$ and $0.420$, and copying would score
below zero; it scores $0.24$ and $0.31$. The floor therefore has to be measured at each horizon that
is evaluated.

Lift is least stable where the floor is highest. With $A$ the model's squared error and
$B = 2S(1-\rho_k)$ the copier's, $\lift = 1 - A/B$ changes by $1/B$ for each unit of error in $A$,
and $B$ vanishes as the floor approaches $1$. This is why a lift of $-4.45$ appears at a floor of
$0.9886$ (Appendix~\ref{app:inhouse}), and it is a reason to report lift with an interval.

\section{Identity predictors and anti-collapse terms}
\label{app:collapse}

Latent prediction objectives admit a degenerate global minimum in which the encoder sends every
observation to the same vector \citep[Prop.~1]{xing2025critiques}, and the usual defences act on the
encoder to prevent
it: variance and covariance penalties \citep{bardes2022vicreg}, action-difference decoding
\citep{deltajepa2026}. A predictor that has learned the identity, $\hat{z}_{t+1} = z_t$, escapes all
of them. It runs on a healthy, full-rank, high-variance embedding, violates no anti-collapse term,
and its latent error looks excellent whenever the floor is high. Anti-collapse terms constrain the
encoder and leave the predictor free, so an identity predictor is caught only by comparing its
prediction with the floor.

\section{The survey: 20 papers}
\label{app:survey}

\input{survey_table.tex}

\end{document}

%% file: table_public.tex
\begin{tabular}{@{}llrrrrrr@{}}
\toprule
& & \multicolumn{4}{c}{Trained checkpoint} & \multicolumn{2}{c}{Random init} \\ \cmidrule(lr){3-6}\cmidrule(l){7-8}
Env & Read-out & Value & Floor & Lift & 95\% interval & Value & Lift \\
\midrule
PointMaze & pooled & 0.8877 & $-$0.3312 & +0.916 & [0.911, 0.920] & $-$118.8873 & $-$89.06 \\
 & per-patch & 0.9969 & 0.9465 & +0.942 & [0.939, 0.945] & $-$0.5715 & $-$28.39 \\
 & per-dim & 0.9991 & 0.9844 & +0.942 & [0.939, 0.945] & 0.5424 & $-$28.39 \\
 & proprio & 0.9884 & 0.5425 & +0.975 & [0.969, 0.980] & 0.3702 & $-$0.38 \\
\midrule
PushT & pooled & 0.8286 & 0.3505 & +0.736 & [0.714, 0.758] & $-$12.4719 & $-$19.74 \\
 & per-patch & 0.9720 & 0.8538 & +0.808 & [0.795, 0.823] & $-$0.2486 & $-$7.54 \\
 & per-dim & 0.9896 & 0.9458 & +0.808 & [0.795, 0.823] & 0.5374 & $-$7.54 \\
 & proprio & 0.9988 & 0.7558 & +0.995 & [0.994, 0.996] & 0.3034 & $-$1.85 \\
\midrule
Wall & pooled & 0.9100 & 0.1564 & +0.893 & [0.875, 0.909] & $-$42.4661 & $-$50.53 \\
 & per-patch & 0.9918 & 0.8738 & +0.935 & [0.926, 0.943] & $-$0.5327 & $-$11.15 \\
 & per-dim & 0.9975 & 0.9615 & +0.935 & [0.926, 0.943] & 0.5328 & $-$11.15 \\
 & proprio & 0.9963 & 0.9790 & +0.824 & [0.789, 0.852] & 0.4484 & $-$25.27 \\
\bottomrule
\end{tabular}

%% file: table_crossmodel.tex
\begin{tabular}{@{}llrrr@{}}
\toprule
Env & Read-out & Zhou DINO-WM & Meta DINO-WM & JEPA-WM \\
\midrule
PointMaze & pooled & $-$0.3312 & $-$0.3656 & $-$0.3870 \\
 & per-patch & 0.9465 & 0.9607 & 0.9650 \\
 & per-dim & 0.9844 & 0.9863 & 0.9877 \\
 & proprio & 0.5425 & 0.5964 & 0.6978 \\
PushT & pooled & 0.3505 & 0.3613 & 0.4011 \\
 & per-patch & 0.8538 & 0.8740 & 0.8681 \\
 & per-dim & 0.9458 & 0.9490 & 0.9491 \\
 & proprio & 0.7558 & 0.8362 & 0.3445 \\
Wall & pooled & 0.1564 & 0.2574 & 0.2592 \\
 & per-patch & 0.8738 & 0.9018 & 0.8887 \\
 & per-dim & 0.9615 & 0.9637 & 0.9595 \\
 & proprio & 0.9790 & 0.9784 & 0.9791 \\
\bottomrule
\end{tabular}

%% file: table_type_a.tex
\begin{tabular}{@{}llrrrrrrr@{}}
\toprule
& & & & Copy & Decoded & \multicolumn{2}{c}{Lift of published} & Lift, ours, \\ \cmidrule(lr){7-8}
Env & Metric & Published & Ours & floor & copy & Lift & 95\% interval & high motion \\
\midrule
PointMaze & LPIPS & -- & 0.0006 & 0.0090 & 0.0094 & -- & -- & +0.94 \\
PointMaze & SSIM & -- & 0.9986 & 0.9949 & 0.9942 & -- & -- & +0.76 \\
PushT & LPIPS & 0.0070 & 0.0067 & 0.0352 & 0.0375 & +0.801 & [0.789, 0.812] & +0.83 \\
PushT & SSIM & 0.9850 & 0.9867 & 0.9654 & 0.9637 & +0.567 & [0.529, 0.600] & +0.68 \\
Wall & LPIPS & 0.0016 & 0.0022 & 0.0236 & 0.0248 & +0.932 & [0.927, 0.937] & +0.94 \\
Wall & SSIM & 0.9970 & 0.9972 & 0.9862 & 0.9855 & +0.782 & [0.768, 0.794] & +0.84 \\
\bottomrule
\end{tabular}

%% file: table_baselines.tex
\begin{tabular}{@{}lrrrr@{}}
\toprule
 & \multicolumn{2}{c}{PushT} & \multicolumn{2}{c}{Wall} \\ \cmidrule(lr){2-3}\cmidrule(lr){4-5}
Method & LPIPS ($\downarrow$) & SSIM ($\uparrow$) & LPIPS ($\downarrow$) & SSIM ($\uparrow$) \\
\midrule
Copy previous frame & 0.0352 (0) & 0.9654 (0) & 0.0236 (0) & 0.9862 (0) \\ \midrule
DINO-WM & 0.007 (+0.80) & 0.985 (+0.57) & 0.0016 (+0.93) & 0.997 (+0.78) \\
R3M & 0.045 (\textbf{$-$0.28}) & 0.956 (\textbf{$-$0.27}) & 0.008 (+0.66) & 0.994 (+0.56) \\
ResNet & 0.063 (\textbf{$-$0.79}) & 0.950 (\textbf{$-$0.44}) & 0.002 (+0.92) & 0.996 (+0.71) \\
DINO CLS & 0.039 (\textbf{$-$0.11}) & 0.973 (+0.22) & 0.004 (+0.83) & 0.996 (+0.71) \\
AVDC & 0.046 (\textbf{$-$0.31}) & 0.959 (\textbf{$-$0.18}) & 0.030 (\textbf{$-$0.27}) & 0.983 (\textbf{$-$0.24}) \\
DINO-WM, 200 trajectories & 0.056 (\textbf{$-$0.59}) & 0.949 (\textbf{$-$0.47}) & -- & -- \\
\bottomrule
\end{tabular}

%% file: table_lewm.tex
\begin{tabular}{@{}llrrr@{}}
\toprule
Arm & Read-out & Value & Floor & Lift \\
\midrule
checkpoint & token & 0.9919 & 0.8103 & +0.96 \\
 & per-dim & 0.9919 & 0.8108 & +0.96 \\
copy-forward & token & 0.8103 & 0.8103 & +0.00 \\
 & per-dim & 0.8108 & 0.8108 & +0.00 \\
random & token & $-$0.0415 & 0.8103 & $-$4.49 \\
 & per-dim & $-$0.0389 & 0.8108 & $-$4.49 \\
\bottomrule
\end{tabular}

%% file: table1.tex
\begin{tabular}{lrrrrr}
\toprule
Read-out & $n$ & below floor & \textbf{passes 0.5 threshold} & floor range & highest below floor \\
\midrule
component-wise & 19 & 14 (74\%) & \textbf{14} & $0.2791$--$0.9962$ & $0.9794/0.9962$ \\
proprioceptive & 98 & 10 (10\%) & \textbf{4} & $-0.1791$--$0.8484$ & $0.8416/0.8484$ \\
full observation & 80 & 15 (19\%) & \textbf{0} & $-0.1971$--$0.7634$ & $0.4272/0.7634$ \\
\midrule
\textbf{Total} & & & \textbf{18} & & \\
\bottomrule
\end{tabular}

%% file: survey_table.tex
One question was put to each paper: when it reports prediction quality, does it also report a trivial
baseline (copy-last-frame, persistence, zero-motion, constant prediction) against which that number
can be read? Verdicts come from the paper text (arXiv HTML full text, or ar5iv where HTML was
unavailable), not from secondary summaries.

\paragraph{Coding criterion.} A paper counts as \emph{reporting prediction quality} if it presents,
at evaluation time, any number or plotted curve measuring agreement between predicted future
representations, or their decodings, and ground truth. Excluded, and coded \texttt{n/a}, are
training-loss curves, task-success metrics, and probes applied to representations encoded directly
from real observations, which involve no prediction step. The last exclusion decides several rows.
The field uses one word, ``probe'', for both a read-out of an
encoded representation and a read-out of a rolled-out prediction, and only the second admits a
persistence floor. \texttt{n/a} is used strictly and is \emph{not} counted as evidence for our claim.
\texttt{partial} means the metric carries an implicit or acknowledged interpretability caveat but no
measured floor.

Of the 14 papers reporting a prediction-quality number, 11 report it with no floor of any kind, 2 are
\texttt{partial}, and 1 measures the floor.

{\footnotesize
\setlength{\extrarowheight}{2.2pt}
\begin{longtable}{@{}>{\raggedright\arraybackslash}p{2.95cm}%
                      >{\raggedright\arraybackslash}p{3.3cm}%
                      >{\raggedright\arraybackslash}p{4.95cm}%
                      >{\raggedright\arraybackslash}p{1.5cm}@{}}
\toprule
\textbf{Paper} & \textbf{Prediction metric reported} & \textbf{Locator / quote} &
\textbf{Verdict} \\
\midrule
\endfirsthead
\toprule
\textbf{Paper} & \textbf{Prediction metric reported} & \textbf{Locator / quote} &
\textbf{Verdict} \\
\midrule
\endhead
JEPA-WM recipe \citep{terver2026drives} & embedding-space error under unrolling; proprioceptive decoding error;
LPIPS of decoded rollouts & \S5.1 lists these as metrics that ``track the world model quality,
independently of the planning procedure''; no trivial predictor is among them &
\texttt{no baseline} \\
DINO-WM \citep{zhou2024dinowm} & LPIPS, SSIM on decoded frames; MSE as planning cost & Tables 5 (LPIPS) and 6 (SSIM). Rows: R3M,
ResNet, DINO CLS, AVDC, ours --- all learned & \texttt{no baseline} \\
On Memory \citep{laird2025memory} & SSIM, LPIPS, image MSE, latent MSE and cycle MSE, $H\in\{10,20,50\}$ &
App. A.3: ``All reported metrics (SSIM, LPIPS, image MSE, latent MSE, cycle MSE) are computed
per-frame against the corresponding ground-truth frame''; rollout is ``fully in imagination''
(\S5.3) from a 9-frame context. Table 1's only reference is a vanilla-ViT baseline of context
length 9 & \texttt{no baseline} \\
stable-worldmodel \citep{bounou2026stableworldmodel} & ``trajectory-level prediction MSE for successful and failed
planning'' & \S4.2 / Fig. 4. \S3.3 baselines are GCBC, GCIQL, GCIVL, DINO-WM, PLDM, LeWM, TD-MPC2;
random policies appear only for \emph{data collection} & \texttt{no baseline} \\
Dreamer 4 \citep{hafner2025dreamer4} & PSNR and SSIM on 16-step generations; FVD & \S4.3 (Fig. 7) reports
PSNR and SSIM against the amount of action-labelled data, scaled between a no-action and an
all-action model; \S4.4 (Table 2) reports FVD for design choices against a naive diffusion-forcing
model; \S3.1's MSE${+}$LPIPS is the tokenizer objective. No trivial-predictor row & \texttt{no baseline} \\
AdaJEPA \citep{adajepa2026} & latent prediction loss plotted per replanning step; no numeric table &
App. C (Figs. 11--14); a latent prediction loss ``such as MSE'' is the adaptation objective. Comparisons are frozen-vs-adapted
variants, adaptation-layer choices, and DINO-WM (Table 2) & \texttt{no baseline} \\
EB-JEPA \citep{ebjepa2026} & Average Precision on Moving MNIST, $k$-step recursive prediction & Fig. 3,
\S5: AP improves with $k$-step recursion, best near $k=4$. Regularizers are compared in Table 1 and
ablated in Table 4; no naive-predictor row appears anywhere & \texttt{no baseline} \\
Jumpy World Models \citep{farebrother2026jumpy} & Earth Mover's Distance for multi-step fidelity & Table 3, \S4.4;
compares td-hc against td-flow and ground-truth policy rollouts & \texttt{no baseline} \\
PERSIST \citep{garcin2026persist} & FVD; 1--5 human ratings & Table 1 (FVD at 200/400/600 frames) and Table 2
(ratings), \S6.1. Baselines are Oasis and WorldMem, both learned generators. The paper's ``persistent'' denotes 3D state, not a persistence
baseline & \texttt{no baseline} \\
Latent Video Prediction \citep{latentvideopred2026} & decoded object-position error of latent
rollouts, $k\in\{1,5,10\}$ steps & App. K.3, Table 18 (``Offline latent rollout quality''): rollout
error beside an oracle that ``decodes the true future latent'', a ceiling; no trivial predictor. The
rest of the paper probes four frozen encoders under ten input perturbations (\S8, App. J) &
\texttt{no baseline} \\
LeWorldModel \citep{lewm2026} & prediction error as a surprise signal on perturbed and unperturbed
trajectories & \S5.2, Fig. 8 (App. F.3): surprise is ``the discrepancy between the model's predicted
future observations and the actual observed future''; perturbed runs are compared with unperturbed
ones, and no trivial predictor appears. Table 1 (\S5.1) probes read \emph{encoded} embeddings &
\texttt{no baseline} \\
IWM \citep{garrido2024iwm} & Mean Reciprocal Rank over a bank of 256 targets & \S4.1. Table 1 includes a
no-conditioning control (MRR $0.00$); chance level is never stated by the paper, and no persistence
or identity floor is reported --- despite identity collapse being the paper's own subject &
\texttt{partial} \\
ATM \citep{atm2026} & action-consistency transfer diagnostics & \S4.5: the score is used
``only for relative ranking rather than as a calibrated predictor of absolute planning success'' &
\texttt{partial} \\
What Can Latent WMs Know? \citep{latentknow2026} & probe $R^2$; force forecast $R^2$ $0.4$\,s ahead &
\textbf{Reports the floor.} Table 16 (real robot, 2,947 training episodes): persistence $0.23/0.31$ against
the model's $0.43/0.47$ (IID/held-out tasks). An untrained network reads current force at
$0.94/0.91$ (Table 15), against the trained model's $0.92/0.88$. A training-mean constant is the
reference for the mass action effect (Table 14). Version 3 also reported a one-step forecast
\citep{latentknow2026v3} & \textbf{\texttt{baseline}} \\
V-JEPA 2 \citep{assran2025vjepa2} & none reported; robot success rates only & Latent $L_1$ prediction losses
appear only as training objectives (eqs. 1--4) and the planning energy (eq. 5); rollout quality is shown qualitatively in App. B.3 &
\texttt{n/a} \\
TD-MPC2 \citep{hansen2024tdmpc2} & none as an evaluation number & The joint-embedding term appears only in the
training objective (Eq. 3); the model is ``implicit, control-centric \ldots\ without decoding
observations'' & \texttt{n/a} \\
SOLD \citep{villar2024sold} & none numerically & \S4.1 assesses 50-frame open-loop rollouts qualitatively
(Fig. 3); Table 1 is final success rates & \texttt{n/a} \\
Causal-JEPA \citep{causaljepa2026} & none directly; CLEVRER VQA and Push-T success & Tables 1--2 (VQA)
and 3 (Push-T); the $L_2$ term in Eq. 5 is training loss only. The OC-JEPA comparison (Table 1,
\S5.1) is an ablation, not a floor & \texttt{n/a} \\
RC-aux \citep{rcaux2026} & none; success rates, a latent reachability diagnostic (\S G.1), and Push-T
probes on representations (\S G.2) & Abstract: a model ``may achieve accurate short-horizon
prediction while still inducing a latent space that is poorly aligned with planning''. The
multi-horizon term is in the Eq. 6 training loss only & \texttt{n/a} \\
VJEPA variational \citep{huang2026vjepavariational} & none; probe $R^2$ for signal recovery on one toy system (\S9) & Compares
linear VAE, pixel-autoregressive, linear JEPA, VJEPA, BJEPA. Proof-of-concept; probe protocol at
App. J.2.4 & \texttt{n/a} \\
\bottomrule
\end{longtable}
}

\paragraph{Limits of this survey.} Verdicts rest on the main paper plus its named appendices. A floor
reported in an unnumbered appendix, a supplementary document, or a code repository would not have
been caught, so the claim is about reporting practice in the paper as read. We also searched
for disconfirming evidence, querying explicitly for ``copy the last frame'', ``persistence baseline'' and
``zero-motion baseline''. The search surfaced \citet{latentknow2026}, recorded above, and confirmed that the practice is
routine in the neighboring fields of motion forecasting and video prediction, so the gap is one of
reporting convention.